\documentclass{IOS-Book-Article} 

\usepackage{mathptmx}
\usepackage{soul}\setuldepth{article}

\usepackage{amsmath,amsfonts}
\usepackage{graphicx}
\usepackage[dvipsnames]{xcolor}
\definecolor{LightGray}{rgb}{0.97,0.97,0.97}
\usepackage{listings} 
\lstdefinelanguage{SPARQL}{
  basicstyle=\small\ttfamily,
  backgroundcolor=\color{LightGray},
  columns=fullflexible,
  breaklines=false,
  sensitive=true,
  frame=bt,
  aboveskip=1em,
  belowskip=1em,
  xleftmargin=.5em,
  xrightmargin=.5em,
  framexleftmargin=.5em,
  framextopmargin=.5em,
  framexbottommargin=.5em,
  framexrightmargin=.5em,
  tabsize = 2,
  showstringspaces=false,
  morecomment=[l][\color{gray}]{\#},      
  morecomment=[n][\color{blue}]{<http}{>},
  morestring=[b][\color{OliveGreen}]{\"}, 
  keywordsprefix=?,
  classoffset=0,
  keywordstyle=\color{Sepia},
  morekeywords={},
  classoffset=1,
  keywordstyle=\color{Purple},
  morekeywords={rdf,rdfs,owl,xsd,purl},
  classoffset=2,
  keywordstyle=\color{MidnightBlue},
  morekeywords={
    SELECT,CONSTRUCT,DESCRIBE,ASK,WHERE,FROM,NAMED,PREFIX,BASE,OPTIONAL,
    FILTER,GRAPH,LIMIT,OFFSET,SERVICE,UNION,EXISTS,NOT,BINDINGS,MINUS,a
  }
}

\usepackage{mathtools}
\usepackage{float}
\usepackage{subcaption}
\usepackage[linesnumbered,ruled,vlined]{algorithm2e}
\usepackage{amsmath,amsfonts,amssymb}

\usepackage{cite}
\usepackage{multirow}

\newcommand{\eat}[1]{}

\def\hb{\hbox to 11.5 cm{}}

\definecolor{neuro_color}{rgb}{0.59, 0.45, 0.65}
\definecolor{symbolic_color}{rgb}{0.51, 0.70, 0.40}

\usepackage{xspace}
\newcommand{\eg}{e.\,g.,\xspace}
\newcommand{\ie}{i.\,e.,\xspace}
\newcommand{\cf}{cf.\xspace}

\newcommand{\mi}[1]{\mathit{#1}}

\newcommand{\dom}{\mathit{dom}}
\newcommand{\aD}{\Omega_{\D}}
\newcommand{\aQ}{\Omega_{A}}
\newcommand{\aF}{\Omega_{\F}}
\newcommand{\av}[2]{#1\text{:}#2}
\newcommand{\F}{{\mathcal F}}
\newcommand{\D}{{\mathcal D}}

\newcommand{\arrowScore}[1]{\stackrel{#1}{\rightarrow}}
\newcommand{\sRule}[3]{#1 \arrowScore{#3} #2}
\newcommand{\aS}{\Omega_{\mi{S}}}

\newcommand{\coloneqq}{\mathrel{\mathop{:}}=}
\newcommand{\DB}{\mathit{DB}}
\newcommand{\true}{\mathit{true}}

\begin{document}

\begin{frontmatter} 
\title{Knowledge-Augmented Explainable and Interpretable Learning for Anomaly Detection and Diagnosis}

\author[A,B]{\fnms{Martin} \snm{Atzmueller}},
\author[B]{\fnms{Tim} \snm{Bohne}}
\author[B]{\fnms{Patricia} \snm{Windler}}
\address[A]{Semantic Information Systems Group, Osnabrück University,\\ Wachsbleiche 27, 49090 Osnabrück, Germany}
\address[B]{German Research Center for Artificial Intelligence (DFKI),\\ Hamburger Str. 24, 49084 Osnabrück, Germany}
\begin{abstract}
Knowledge-augmented learning enables the combination of knowledge-based and data-driven approaches. For anomaly detection and diagnosis, understandability is typically an important factor, especially in high-risk areas. Therefore, explainability and interpretability are also major criteria in such contexts. This chapter focuses on knowledge-augmented explainable and interpretable learning to enhance understandability, transparency and ultimately computational sensemaking. We exemplify different approaches and methods in the domains of anomaly detection and diagnosis -- from comparatively simple interpretable methods towards more advanced neuro-symbolic approaches.
\end{abstract}


\end{frontmatter}

\section{Introduction}

In the context of complex systems, anomaly detection~\cite{chandola2009anomaly,pang2021deep} and diagnosis~\cite{szolovits1988artificial,Pup:93,trave2014bridging,liu2018artificial,su2024machine} are important tasks for gaining insight into the behavior of a system, for example to prevent critical situations or to address specific problems
like faults or failures in technical systems. However, this does not only apply to technical systems~\cite{majstorovic1990expert,reinertsen1996residual,lei2020applications,chen2012technical}, but also to biological systems~\cite{ishiguro1994fault,alcaraz2012interval,fezai2021fault} or the medical domain~\cite{EO:00,BEFKP:02,PABHLB:08}. Here, with the ever-growing amount of collected data, we can
investigate, analyze and interpret observed phenomena at considerably larger scale and in much more detail.

Particularly in prominent areas of artificial intelligence and machine learning, such as anomaly detection and diagnosis, there usually exists some knowledge which can be exploited for guiding and/or augmenting specific
learning methods. Augmentation can take place at different levels, \eg serving as a starting point for these models, providing constraints for them, or integrating specific hybrid models consisting of both data-driven
as well as knowledge-based components. However, with the constantly growing amount of available data and accordingly refined models, \eg from the field of machine learning, gaining insights into a system, model -- or a decision made
by those -- becomes much more difficult due to its inherent complexity~\cite{rudin2019stop}.

Hence, in many domains and applications an important goal is to enable human insight into the methods and/or models in order to ultimately allow for computational
sensemaking~\cite{Atzmueller:18:Declare}. In such cases, explainability and interpretability play a major role for making sense of the according systems, models and methods.
In general, explainability and specifically the term explanation~\cite{Roth-BerghoferRichter2008a,AR:10a} have been widely investigated in different disciplines. Providing answers to questions, it should support humans in their decision-making~\cite{Schank86a}. This is particularly relevant for complicated black box models being applied in sensitive application contexts like medicine, Industry 4.0, etc. (\cf~\cite{rudin2019stop}).
While interpretability~\cite{biran2017explanation,molnar2020interpretable,das2020taxonomy,VAT:21,rudin2022interpretable} and explainability~\cite{gunning2017explainable,barredo2019explainable,burkart2021survey} are often used synonymously, they can be distinguished in the following way: ``Systems are interpretable if their operations can be understood by a human, either through introspection or through a produced explanation'' \cite{biran2017explanation}, whereas ``explainability is viewed as an active characteristic of a model or method generating an explanation'' \cite{barredo2019explainable}. Thus, while interpretable methods are always explainable, explainable methods are not always interpretable, specifically if  the inner mechanics of the system are not understandable by a human.
In particular, rule-based and pattern-based methods, \eg~\cite{Atzmueller:15a,furnkranz2020cognitive,furnkranz2015brief,molnar2020interpretable,roscher2020explainable}, have shown considerable impact and promise in this respect, since such models are directly interpretable and thus also explainable. In contrast, black box methods such as deep learning approaches typically require post-hoc methods, \eg~\cite{ribeiro2016should,lundberg2017,Ribeiro2018AnchorsHM,gradcam2016}, for generating explanations, \eg~\cite{han2022explanation,iferroudjene2023methods,AFKS:24}.
For the knowledge-augmented interpretable and explainable dimensions, there are a variety of methods that can be applied, \eg ranging from simple (linear) models to more complex but interpretable pattern-based methods, and finally to deep-learning-based approaches.
In the latter case, this specifically refers to neuro-symbolic approaches~\cite{garcez2022neural,hitzler2022neuro, garcez2022neural,liu2021machinery} whenever advanced deep learning methods are complemented by symbolic components, \eg domain knowledge. This knowledge is typically provided via ontologies or knowledge graphs~\cite{hogan2021knowledge,ji2021survey}, but can also refer to other forms. Thus, with such neuro-symbolic approaches it is possible to integrate rich data-driven methods with according knowledge-based representations, \eg applying domain knowledge.

This chapter aims to provide a focused view on these topics by reviewing and discussing exemplary methods in this domain, specifically considering the articles~\cite{AS:17,SA:18} for a targeted view on pattern mining in the context of anomaly detection, as well as~\cite{ABP:05Score,BAKP:06,ABP:06} for learning scoring systems in the form of diagnostic scores, and a significantly adapted summary of the article~\cite{bohne:2023}, covering a neuro-symbolic approach for anomaly detection and diagnosis. Thus, we review simple knowledge representations such as diagnostic scores as well as interpretable pattern-based methods in combination with different forms of symbolic knowledge. Furthermore, we also look into a hybrid neuro-symbolic method integrating deep learning with symbolic knowledge.

The remainder of this chapter is structured as follows: Section~\ref{sec:pattern:mining} summarizes knowledge-augmented pattern mining and its application for anomaly detection.
Next, Section~\ref{sec:scoring:systems} briefly reviews scoring systems, how to learn and refine them in a knowledge-augmented way, as well as an exemplary application in the medical domain.
After that, Section~\ref{sec:explainable:diagnosis} summarizes a neuro-symbolic approach for explainable diagnosis, exemplified in a technical diagnosis domain.
Finally, Section~\ref{sec:conclusions} concludes with a summary and presents interesting directions for future research.

\section{Knowledge-Augmented Pattern Mining for Anomaly Detection}\label{sec:pattern:mining}

In general, pattern mining provides a broadly applicable and powerful set of methods for exploratory data analysis, knowledge discovery, and computational sensemaking on complex data~\cite{AP:05,ABP:06,Atzmueller:15a,Atzmueller:18:Declare,fournier2022pattern}. Exemplary methods include association rule mining~\cite{AS:94}, subgroup discovery~\cite{Wrobel:97,Atzmueller:15a}, exceptional model mining~\cite{LBA:12,Atzmueller:15a,Duivesteijn:2016:EMM:2877058.2877103} as well as graph (pattern) mining~\cite{kaytoue2017exceptional,ASSB:19,atzmueller2021mining}. In particular, whenever semi-automatic and interactive approaches can be provided, constraints and user interests can often be directly integrated~\cite{AP:08,AL:12a,dzyuba2013interactive}. An alternative and/or complementary approach is to inject domain knowledge into the pattern mining process. Here, first approaches for such an integration utilizing knowledge graphs -- exploiting information formalized in ontologies and a set of instance data -- have been proposed in the area of semantic data mining~\cite{VPL:14,rauch2014learning,dou2015semantic,nalepa2021semantic}.
Furthermore, \cite{AS:17} presents a mixed-initiative approach for semantic feature engineering using a knowledge graph, which is constructed in a semi-automatic process. Here, the resulting features are then provided for data analysis. A similar approach is applied in~\cite{FEE:ATP:2016}. Here, data from heterogeneous data sources is integrated into a knowledge graph, which then provides the basis for knowledge discovery.

In this chapter, we review and discuss approaches and an application use case for knowledge-augmented pattern mining~\cite{Atzmueller:15a,AS:17,SA:18}, specifically focusing on subgroup discovery connected to domain knowledge formalized in knowledge graphs. Hence, we focus on an integrated approach, exploiting knowledge-based semantic structures, \ie sets of knowledge components connected to a local pattern mining method. Exceptions and anomalies can then be detected via appropriate measures of interestingness.

\subsection{Pattern Mining Using Subgroup Discovery}

Subgroup discovery is a flexible approach for exploratory data analysis and knowledge discovery, enabling computational sensemaking on complex data. In its simplest form, subgroup discovery aims to detect relations between dependent (characterizing) variables and an independent target concept, \eg comparing the share or the mean of a nominal/numeric target variable in the subgroup vs. the share or mean in the total population, respectively~\cite{Wrobel:97,Atzmueller:15a,LAP:16}. Thus, in the simplest case, a binary target variable is considered, where the share in a subgroup can be compared to the share in the dataset in order to detect (exceptional) deviations. In the most general formalization, however, the interestingness of a pattern is flexibly defined by a specific interestingness measure. For example, the so-called target concept can then also relate to a complex model, \eg a linear regression model or Bayesian network induced on the subgroup vs. one such model learned from the total dataset, where significant deviations can be estimated.
The latter case is the focus of exceptional model mining~\cite{LBA:12,Atzmueller:15a,Duivesteijn:2016:EMM:2877058.2877103} as a specialization of the general subgroup discovery approach. The patterns are typically described by a description language using conjunctions of attribute-value restrictions, \ie referring to value constraints on features, similar to simple rules~\cite{Atzmueller:15a}. In contrast to typical machine learning approaches, subgroup discovery aims at discovering local patterns, \eg ``nuggets'' in the data~\cite{Kloesgen:02a,knobbe2008local}. Due to their simple conjunctive structure, subgroup patterns are typically interpretable, relating to interpretable machine learning~ \cite{biran2017explanation} which is particularly important when the discovered patterns need to be actionable~\cite{LCGF:04,APB:05b,antonacopoulou2007actionable}.

\subsection{Subgroup Discovery Basics}

For a more formal view on subgroup discovery, we adapt the presentation in~\cite{Atzmueller:15a}, to which we refer to for a detailed discussion. Basically, we distinguish between \emph{patterns} -- which are typically provided in conjunctive form -- and \emph{subgroups} as extensions of such descriptive patterns. The patterns cover specific subsets of a dataset (or database). Then, an \emph{interestingness measure} (also called \emph{quality function}) formalizes which patterns are considered interesting for ranking patterns (and subgroups). For the interestingness measure, this typically includes the concept of interest, as already sketched above. As we will see below, such an interestingness measure can also capture a typical \textit{key performance indicator} (KPI), \eg for process analysis, which formalizes the specific constraints of the user relating to discovering interesting subgroups for the analysis task at hand.

More formally, we consider a database $\DB=(I, A)$, which is given by a set of individuals $I$ (often called instances) and a set of attributes $A$. In the following, we restrict our focus to nominal attributes, where numeric attributes can be handled via discretization, \eg~\cite{Fayyad:1993gd,LAP:16}. For nominal attributes, a selection descriptor $(a=v)$ is a Boolean function $I \rightarrow \{0,1\}$ that is true if the value of attribute $a \in A$ is equal to $v$ for the respective individual. $\Delta$ denotes the set of all selection descriptors. As basic elements in subgroup discovery, we consider patterns, \ie subgroup descriptions and subgroups.
A subgroup description $p$ is given by a set of selection descriptors $p = \{d_1, \ldots, d_l\}, d_i \in \Delta, i = 1, \ldots, l$, \ie having length $l$, which is being interpreted as a conjunction of the contained descriptors. We also call $p$ a (complex) pattern.
In principle, such a pattern can be interpreted as the body of a conjunctive rule. Furthermore, we distinguish between the pattern -- as an intensional description -- and the extent of the pattern, \ie the respective subgroup described by the pattern.
A subgroup
\(s = \{i\in I \,|\, p(i) = \true\}\), is the set of all individuals that are covered by the subgroup description $p$.

In a top-$k$ setting, a subgroup discovery algorithm returns the top-$k$ subgroups according to a given quality function $q \colon 2^\Delta \rightarrow \mathbb{R}\,,$ \cf~\cite{Atzmueller:15a}. The applied quality function thus estimates the interestingness of a pattern. For example, for a binary target concept, a standard quality function combines the size $n_p = |s|$
of a subgroup $s$ described by pattern $p$, \ie its support, and the share $t_p$ of the target concept $t$ in $s$, \ie its confidence as follows: $q^e(p) = n_p^e \cdot (t_p - t_0)$, where $t_0$ denotes the (default) share of the target concept in the database $\DB$, and $e \in \mathbb{R}$. Standard quality functions include the so-called \textit{Piatetsky-Shapiro} quality function $q^1$, the binomial test quality function $q^{0.5}$, and the gain quality function $q^0$. The latter focuses on improving the target share and thus can be applied for finding patterns with a high confidence, however, it then requires a minimal size threshold ensuring sufficient support.
While a quality function provides a ranking of the discovered subgroup patterns, often also a statistical assessment of the patterns is useful in data exploration and knowledge discovery. Quality functions that directly apply a statistical test, for example the \textit{Chi-square} quality function, \eg~\cite{Atzmueller:15a}, provide a $p$-value for simple interpretation.

Using a given subgroup discovery algorithm, \eg~\cite{Atzmueller:15a} the result of top-$k$ subgroup discovery is then the set of the $k$ patterns $p_1, \ldots, p_k\,,$ where $p_i \in 2^\Delta$, with the highest interestingness value according to the applied quality function.

\subsection{Use Case: Anomaly Detection in Industrial Logistics Data}

In the following, we summarize an application use case of knowledge-augmented pattern mining for anomaly detection in the domain of industrial logistics, adapting the presentation in~\cite{SA:18}.
Here, the goal for anomaly detection is to identify exceptional patterns in the context of inventory differences.
More specifically, the major goal of the analysis was to identify specific logistic processes in a larger production context which could potentially indicate erroneous financial assessments.
These are called inventory differences in this context, \cf~\cite{SA:18}.
Overall, the whole process was considerably assisted by domain specialists, where large-scale data as well as domain dependencies were integrated into a knowledge graph representation.
A further important goal was to provide interpretable representations in order to enable explanation and transparency, which was addressed via a knowledge-augmented pattern mining approach using subgroup discovery.

\begin{figure}[htb] 
   \centering 
 \includegraphics[width=.99\textwidth]{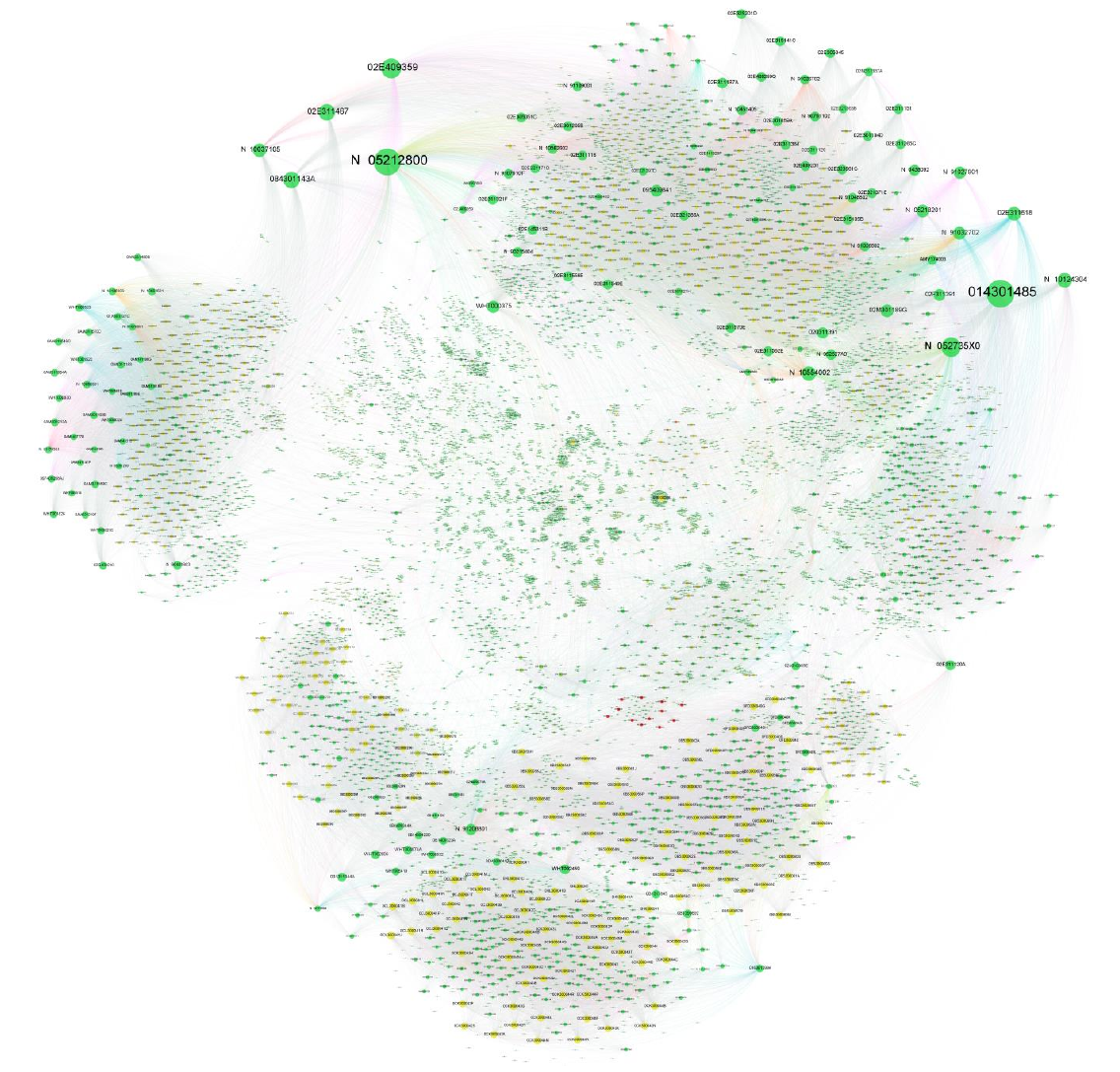} 
   \caption{Visualization~\cite{SA:18} of the  structure data graph (bill of materials information). Each node represents some material, with a node's size proportional to its degree. The color of a node tends to indicate its price in a traffic light scheme from green to yellow and red, where more inexpensive (basic) materials are colored in green and more expensive materials  range from yellow to red. We refer to~\cite{SA:18} for a more detailed discussion.} 
   \label{fig:gephi-graph-vis} 
 \end{figure}

Initially, business and data understanding as in the CRISP-DM model~\cite{CRISP00} for data mining was applied in order to determine the most relevant data sources for the targeted inventory difference problem. This resulted in a rather complex structure of different databases and further process information (see~\cite{SA:18} for more details) as the basis for building the knowledge graphs. Here, two graphs were considered: the so-called \textit{structure data graph} (see Figure~\ref{fig:gephi-graph-vis} for a visualization) that represents information and dependencies of the specific bills of materials, and the \textit{accounting data graph} which captures dependencies of material flow information within the respective production processes.

For anomaly detection and identifying relevant patterns, a key challenge was to assess the relevant bookings of a material and its dependent materials. Given the complexity and volume of data, manual analysis was only feasible for small cases. Therefore, for pattern mining, a specific \textit{key performance indicator} (KPI) was developed as a complex interestingness measure. This KPI was applied as a data analytics feature for each material node in the structure data graph.
Essentially, this KPI operates across both graphs: The structure data graph, which represents the bill of materials (\ie the part--part structure), and the accounting data graph, which records bookings (\ie information about movements of parts). Basically, the KPI traces the construction dependencies within the structure data graph, from basic materials (\eg screws, aluminum) to final products (\eg gears, exhaust systems, etc.) that incorporate those materials. For each material node, the accounting data graph provides the corresponding booking information and effective quantities.
The domain specialists expected that the aggregated amounts from the two graphs would cancel each other out if the booking calculations and movements were correct. This method essentially provided a kind of logistical balance for each material node of the graph. If the respective balance aligned with the expectations of the domain specialists, \ie that the data, calculations and expected movement information was correct and reflected their understanding of the respective processes, then the KPI would be close to zero. Any deviation from zero would therefore indicate potential issues in the data, \ie a divergence from zero would indicate either data discrepancies or issues in the domain model. Consequently, anomalous subgroups exhibiting such deviations were identified and investigated.

The analysis revealed significant deviations from the experts’ expectations. For example, when analyzing subgroups for different work shifts, it was discovered that specific types of logistic bookings were strongly correlated with particular shifts. Additionally, when focusing on material application and price fluctuations, technical issues were identified in a data import process. This problem was flagged by patterns showing bookings with empty storage groups, which conflicted with the experts’ expectations. Furthermore, subgroup analyses targeting the fluctuations of component prices revealed incomplete price data for a large group of parts. In subsequent iterations, another systematic issue was uncovered, for example, where cost center IDs for the same physical cost center were inconsistent across different but interrelated logistic systems. We refer to~\cite{SA:18} for a more detailed discussion. Potential extensions of such knowledge-augmented analysis method can consider the graph structures directly, for detecting according graph patterns, \eg~\cite{kaytoue2017exceptional,AK:18,ABK:19}, also when integrating domain knowledge, \eg~\cite{FEE:ATP:2016}.

\section{Knowledge-Augmented Interpretable Learning of Diagnostic Scoring Systems}\label{sec:scoring:systems}

Knowledge-based diagnosis is a prominent area of research, including early approaches in the field of expert systems and knowledge systems. These often use explicitly formalized knowledge, often complemented by ontology-based approaches~\cite{HWL:83,Pup:93,puppe1998knowledge,SFDB:99,baumeister2015knowledge,chi2022knowledge,nan2008real,li2020process,chi2022knowledge,stefik2014introduction,steenwinckel2018towards} as well as more recent data-driven learning methods~\cite{Kononenko:01,klemettinen1997data,leitich2002evaluation,ABP:05Score,ustun2016supersparse,ustun2019learning,yun2021knowledge}.
These become more and more important in the context of the diagnosis of complex systems that generate large amounts of data, \eg in medical~\cite{Kononenko:01,PBAHB:05,PABHLB:08} or industrial~\cite{gao2015survey,FEE:ATP:2016,chi2022knowledge} domains.
In general, the automated diagnosis of such complex systems is a challenging task. To successfully apply either approach, adequate knowledge and data is required, while the connection between the systems, \eg for handling different types of information in respective abstractions, is not automatically provided. Furthermore, knowledge acquisition is often costly, while purely data-driven techniques require large amounts of data of sufficient quality. Therefore, hybrid knowledge-augmented approaches can combine the advantages of both purely knowledge-based and solely data-driven approaches, \ie an initial set of high-quality knowledge, which can be extended, refined and fine-tuned by data-driven learning. In this section, we summarize such an approach~\cite{ABP:05Score,BAKP:06,ABP:06} based on scoring systems~\cite{Pup:00,ustun2019learning}, namely diagnostic scores~\cite{Pup:00,EO:00}. We discuss initial knowledge-augmented learning as well as knowledge refinement, and present an exemplary use case in the medical domain.

\subsection{An Overview on Scoring Systems}

For diagnosis, \emph{diagnostic scores} are a widely used formalism for medical decision-making, \eg \cite{Pop:82,MPM:82,EO:00,BEFKP:02}. We follow the presentation in~\cite{ABP:05Score} for such \emph{scoring systems}, also including refinement and adaptation methods described in~\cite{BAKP:06,ABP:06}. In particular, below, the term \emph{scoring system} denotes a set of diagnostic scores in the form of \emph{scoring rules}, \ie the resulting \emph{(scoring) rule base}.

Essentially, scoring systems as linear models are a relatively simple and lightweight knowledge representation.
For a diagnostic application, we distinguish inputs (\emph{features}) and outputs as \emph{concepts} (\emph{diagnoses}) of a model.
In order to derive a concept (\eg a diagnosis), a limited number of features is utilized in a regular and easy-to-interpret manner. Then, in its simplest form, each observed feature contributes a single point to a \emph{score}. If the score exceeds a predetermined threshold, then the diagnostic concept is considered as established. There are variations where multiple categories are used instead of just using one point, where both negative and positive contributions are considered, and where multiple thresholds are employed to express varying degrees, such as distinguishing between ``possible'' and ``probable'' in the derivation of a concept. Furthermore, we can also generalize categories to arbitrary points (natural numbers) or real values for general scoring systems. For diagnostic scores, however, simple to interpret \emph{(symbolic) categories} are often preferred for their improved understandability.
Diagnostic scores are typically implemented using scoring rules, which assign specific points to a diagnosis as defined in the rule action. Unlike general rules, scoring rules usually do not involve logical combinations in their preconditions.
However, they can be arranged hierarchically such that a concept inferred through a score can be used to infer another concept, enabling more complex (hierarchical) models.

\subsection{Diagnostic Scores}

In the following, we provide a more formal view on diagnostic scores, summarizing the presentation and definitions given in~\cite{ABP:05Score}.
For a diagnostic model implemented as a knowledge system, we consider attributes respectively attribute values (or findings) as inputs, and diagnoses as its outputs.
We define $\aQ$ as the universal set of all attributes available in the problem domain. Then, a value $v \in \dom(a)$ assigned to an attribute $a \in \aQ$ is called an \emph{attribute value} or \emph{finding}. Furthermore, let $\aF$ denote the set of all possible attribute values in the given problem domain. An attribute value $f \in \aF$ is denoted by $\av{a}{v}$ for $a \in \aQ$ and $v \in \dom(a)$.
The set $F_a \subseteq \aF$ of possible attribute values for a given attribute $a$ is
defined as $F_a := \{f \in \aF\,|\, f = \av{a}{v} \wedge v \in dom(a)\}$.
Each attribute value $f \in \aF$ is defined as a possible input of the model, \eg a
diagnostic knowledge system.
We denote a \emph{diagnosis} $d$ as a possible output of the respective model.
$\aD$ is defined as the universe of all possible diagnoses for a
given problem domain.

For modeling scoring systems (or diagnostic scores) we consider simple scoring rules.
A \emph{simple scoring rule} $r$ is denoted as
\(
    r := \sRule{f}{d}{s} \, ,
\)
where $f \in \aF$ is an attribute value, $d \in \aD$ is the targeted
diagnosis, and $s \in \aS$ denotes a symbolic confirmation category. $\aS$ denotes the full set of symbolic confirmation categories. Basically, enlarging this set enables a more fine-grained approach for learning and fine-tuning, while a coarser approach facilitates manual specification, fine-tuning and adaptation, since the choice of the specific categories is restricted. In general, the category $s$ reflects the concept of the points in a scoring system in a symbolic and thus more interpretable form for the given rule $r$.
Diagnostic scores are then used to represent a qualitative approach for deriving diagnoses with symbolic confirmation categories.
These categories state the degree of confirmation or disconfirmation of a particular diagnosis.  Categories can be aggregated according to some suitable aggregation scheme, \eg such that four equal categories together result in the next higher category. Then, according to the obtained confirmation categories (the symbolic score), the respective concept (diagnosis) is established.

\subsection{Learning Diagnostic Scores}

The approach described in~\cite{ABP:05Score} aims to learn diagnostic scoring rules in a data-driven way by estimating the association strength between all findings $f \in \aF$ and all diagnoses $d \in \aD$, basically relying on statistical tests. In principle, first so-called diagnostic profiles are constructed for each diagnosis $d \in \aD$ for which all contributing findings $f \in \aF$ having a positive or negative association with the diagnosis above a certain user-specified threshold are considered. Then, the respective score of the rule, \ie the symbolic scoring confirmation category is estimated based on the predictive performance of the finding with respect to the diagnosis. For this, standard measures~\cite{10.1007/978-3-031-35314-7_2,rainio2024evaluation} like precision and the false positive rate are applied, \cf~\cite{ABP:05Score} for details. The direction of the statistical association is used for inferring positive or negative scores (points), \ie for a positive association a positive score is applied, and for a negative association a negative score likewise. Finally, the derivation of the symbolic confirmation categories is obtained via a mapping table, which is predefined according to domain knowledge and user requirements, \eg regarding the applied thresholds for determining the categories as well as the number of categories. If being integrated with other optimization/refinement approaches (\eg as discussed below), then this mapping can also be applied at a later stage, for example, when optimization/refinement has been finalized.

Since usually a large number of (statistical) associations are derived, it is important to prune the resulting score rule base. This can be done via domain knowledge, \eg such that only associations between \emph{abnormal} findings and diagnoses are considered. Also, in the medical domain it is often possible to partition the set of findings and diagnoses into partially disjunct subsets, \ie \emph{partition classes}, corresponding to specific problem areas of the application domain, \cf~\cite{ABP:03}. In the medical domain of sonography, \eg these subsets can correspond to problem areas like \emph{liver}, \emph{pancreas}, \emph{kidney}, \emph{stomach} and \emph{intestine}. Then, for learning associations between findings and diagnoses, these associations can be restricted to certain of these subsets, such that only rules based on associations within a specific subset are derived. Furthermore, we can apply heuristic pruning, \eg to remove rules if they contribute only negative but no positive scores for a diagnosis, taking into account whether a diagnosis can be (de-)established or not based on the available rules. For more details, we refer to~\cite{ABP:05Score}.
We briefly revisit the results of~\cite{ABP:05Score} as an example enabling the bootstrapping of diagnosis systems below.

In addition to the simple scheme outlined above, there are also extensions of such a simple rule-based diagnostic score scheme so that more complex rules can be learned, \eg using subgroup discovery, \cf~\cite{ABHRP:05,ABKP:07}. Then, either the simple learning approach or the more complex one -- for obtaining more complex rules, \ie rules with more than one element in the precondition -- can be applied for rapid knowledge capture. In particular, these can be applied in a knowledge-augmented way, \eg incorporating domain knowledge for learning the rules as well as extending a given model (in the form of a rule base) with new rules.

\subsection{Refining Scoring Rules}

Whenever an initial scoring rule base has been provided, either manually or using a learning approach, its optimization and maintenance are important for the further development of the according scoring system~\cite{ABHRP:05,BAKP:06,ABP:06,ustun2016supersparse,ustun2019learning}. For example, in the medical domain, scoring systems are often constructed manually by domain specialists first.
Here, a domain specialist typically assigns ratings to all correlations between findings and solutions, estimating the respective point score based on the domain specialist's expertise. However, if the impact of a combination of attribute values is not proportionately strong compared to the effect of individual attribute observations, for example, then this knowledge representation becomes inadequate. This is because the contribution of attribute values to the diagnostic score is strictly linear. Then, refinement and adaptation of the scoring rules contributions becomes necessary, which can be supported by automatic methods. This includes semi-automatic methods~\cite{ABHRP:05,BAKP:06,ABP:06} as well as optimization-based approaches~\cite{ustun2016supersparse,ustun2019learning}.

For refinements of scoring systems, providing control during the modification and refinement process is usually important. This is enabled, for example, via interactive approaches, \eg~\cite{ABHRP:05}, based on pattern mining methods using subgroup discovery for detecting patterns that cause incorrect behavior of the scoring rule base such that errors in the predictions are identified. For example, rules can be adapted and/or extended, \eg by adding conditions to the rule, by modifying the respective action, or by tweaking the assigned confirmation category. Using visualization, the respective analytical elements uncovered by the pattern mining approach can be interactively checked, and according modifications can be performed, \cf~\cite{ABHRP:05} for further details.

Furthermore, we can apply semi-automatic optimization methods~\cite{BAKP:06,ABP:06} for getting proposals for adapting the respective confirmation categories based on ideas of perceptron learning. Given suitable test cases for the scoring systems, adaptations of the points / confirmation categories can then be proposed -- similar to adapting the perceptron weights during its training phase, \cf~\cite{ABKP:07}. There are also further optimization approaches~\cite{ustun2016supersparse,ustun2019learning} both for learning and refining scoring systems. Combinations of the aforementioned techniques are also possible so that interactive approaches can be linked with advanced optimization and refinement techniques, as well as integrated with other diagnosis formalisms, \eg~\cite{papakonstantinou2020integra}.

\subsection{Use Case: Bootstrapping Diagnostic Knowledge Systems}

A prominent use case is to bootstrap diagnostic knowledge systems, that is, to support the domain
specialists when building such a diagnostic knowledge system from scratch. Applying the learning method, an initial set of scoring rules can be obtained, which can be adapted and refined in a second step.

\begin{table}[htb]
\caption{Exemplary evaluation results~\cite{ABHRP:05}: Options for including domain knowledge (partition class information, abnormality information) and heuristic pruning (\cf~\cite{ABHRP:05} for details). $\varnothing$Rules indicates the average number of rules per diagnosis, $\varnothing$AV the average number of attribute values used, $\varnothing$SC the average number of score categories, and $\varnothing$ACC the mean accuracy.}\label{tab:score:learning:results}
\begin{center}
\begin{tabular}
{ c | c | c | c | c | c }
Partition Class / Abnormality / Pruning & $\varnothing$Rules & $\varnothing$Size & $\varnothing$AV & $\varnothing$SC & $\varnothing$ACC \\
\hline
$-$ / $-$ / $-$ & 10.93$\pm$5.18 & 786.80 & 245.80 & 2.69 & 0.90\\
+ / + / $-$ & 3.37$\pm$1.45 & 242.90 & 112.60 & 1.77 & 0.85\\
+ / + / + & 2.12$\pm$0.96 & 152.70 & 82.50 & 0.92 & 0.85\\
\end{tabular}
\end{center}

\end{table}

In~\cite{ABHRP:05}, the presented method was evaluated in such a context for learning an initial scoring system in the domain of medical diagnosis, using a case base collected by the fielded knowledge system \textsc{SonoConsult}~\cite{HBMSPB:04,PABHLB:08}: It targets the field of abdominal ultrasound (liver, portal tract, gallbladder, spleen, kidneys, adrenal glands, pancreas, stomach, intestine, lymph nodes, abdominal aorta, cava inferior, prostate, and urinary bladder) mainly for structured documentation and diagnosis.
In the evaluation, 1340 instances from \textsc{SonoConsult} were applied, with an average number of $4.32\pm2.79$ 
diagnoses and an average number of $76.89\pm20.59$ relevant attribute values per instance, \cf~\cite{ABHRP:05}. For the experiments, the presented method for learning diagnostic scores was applied in a standard $10$-fold cross-validation evaluation setting. Table~\ref{tab:score:learning:results} summarizes the main results for the correlation threshold of $0.5$. Specifically, we observe that applying domain knowledge can significantly reduce the number of learned rules, so that the understandability of the rule base is improved. However, the accuracy is also reduced slightly when rules are pruned via applying domain knowledge. With additional heuristic pruning of the scoring rules -- \eg by removing non-contributing rules as discussed above -- the rule base is further reduced so that only the ``more meaningful'' rules remain. Thus, the presented method allows for a fine-grained approach for learning scoring systems, where a knowledge-augmented approach can remove potential spurious associations which can cause overfitting of the learned scoring rule base, while also reducing the total number of learned rules. The latter then improves interpretability and understandability of the resulting scoring system, which can then be extended and refined further.

\section{Explainable Neuro-Symbolic Anomaly Detection and Diagnosis}\label{sec:explainable:diagnosis}

For the explainable neuro-symbolic approach to anomaly detection and diagnosis described in~\cite{bohne:2023}, the overall idea is to utilize a \emph{knowledge graph} (KG) that guides the diagnostic process, combined with \emph{neural networks} that enable the interpretation of sensor signals suggested by the KG for investigation.
Essentially, this integrates knowledge- and machine-learning-based fault diagnosis, combining both paradigms. A key element is an iterative diagnosis cycle in which rough hypotheses are refined using both knowledge-based and data-driven methods. Explainability is essential for diagnosis and is enabled via the hybrid approach.
The advantage of the neuro-symbolic approach to the problem of automated diagnosis compared to previous methods is that it is designed in a way that both paradigms are mutually beneficial. Thus, it is motivated by the previous lack of explainability and exploitation of available domain knowledge in data-driven methods, and the extensive manual effort and shortcomings with respect to sensor signal evaluation in expert systems.
Ultimately, a comprehensive explanation is constructed by contextualizing all diagnostic artifacts with symbolic state transitions as an explanatory report.
Additionally, they augment the KG and enable to learn the most significant aspects of the signal types over time.

Below, we first summarize how to generate explanations for time series predictions using saliency maps~\cite{ullah2020brief,he2022survey}. After that, we review neuro-symbolic fault diagnosis as presented in~\cite{bohne:2023} before we summarize and discuss an exemplary application use case in the automotive domain.

\subsection{Saliency Map Generation for Time Series \& Explanatory Reports}
\label{sec:cam}

Despite the commonplace of a black box nature of deep learning approaches,
CNNs, among other architectures, offer the advantage of explainability, \eg through \emph{Class Activation Mapping}
techniques \cite{Selvaraju:2017,he2022survey}, providing explanatory insights into the temporal / spatial segments that are important for a network's prediction, \cf~\cite{ullah2020brief,he2022survey}.
A crucial idea of the overall approach in \cite{bohne:2023} is to close the loop and feed this information back into the KG. In case of an anomaly, it is the information
where the error is located in the signal, \ie generally where the system tells us to look to identify the problem under consideration, the region of interest (ROI).
This is very valuable knowledge that is unavailable a priori.
Moreover, it is a very useful debugging resource, highlighting issues such as overfitting and deviation from expert judgements.
Thus, subsequent to the classification of a signal, the explanation of the decision proceeds on the basis of \emph{Class Activation Maps} (CAMs). This is to ensure that
not only accurate predictions are obtained, but also predictions that are comprehensible for users, which should reduce the proneness to errors, and can further enable
computational sensemaking
\cite{holzinger2013human,Atzmueller:18:Declare}.

There are several techniques used in deep learning to visualize areas of an image that are most relevant to predicting a certain class, \eg \emph{Grad-CAM}
\cite{Selvaraju:2017}, \emph{HiResCAM} \cite{Draelos:2021}, \emph{Grad-CAM++} \cite{Chattopadhyay:2018},
\emph{Score-CAM} \cite{Wang:2020}, \emph{SmoothGrad} \cite{Smilkov:2017}, and \emph{LayerCAM} \cite{Jiang:2021}. They provide a way to interpret the decision
made by an ANN model (with compatible architecture, unless model-agnostic) by highlighting the regions of the input image that contribute the most to the classification result. Then, this can then provide important information for human interpretation of the model and according explainability for a prediction, towards computational sensemaking~\cite{lin2021you,kaur2022sensible}.

The basic idea
behind vanilla \emph{Grad-CAM} is to use the gradients of the output class with respect to the feature maps of the last convolutional layer in the network~\cite{Selvaraju:2017}.
The result is a set of gradient feature maps with the same dimensions as the feature maps of the last convolutional layer. Each element represents the
sensitivity of the output score with respect to the corresponding element in the feature map, \ie how much the output would change if the corresponding element in the feature map
were to change. The gradient feature maps are used to compute an importance weight for each feature map in the final convolutional layer w.r.t. the considered prediction. This is done via \emph{Global Average Pooling} (GAP). Finally, the feature maps are weighted with these values and then added up to generate a heatmap that can be
overlaid on top of the original image to provide a visual explanation. The regions with higher values in the heatmap correspond to the regions of the image that had the greatest impact on the model's prediction. Based on the considered architecture, the heatmap may need to be rescaled to match the original dimensions of the input, which may impair
its accuracy.

\emph{Grad-CAM} is primarily designed to work with CNNs, since it relies on the feature maps of the last convolutional layer.
However, the general idea of using gradients to visualize the importance of input features can be applied to other architectures as well. The technique is based on the idea of a
hierarchy of feature representations, as in CNNs, where the output of each layer can be interpreted as a set of learned features that capture increasingly complex aspects of
the input. The choice of the heatmap generation technique depends on the specific application and requirements of the task.
Open source implementations of some of these techniques are available as part of the \textit{tf-keras-vis}\footnote{https://keisen.github.io/tf-keras-vis-docs/} library,
which is primarily intended to be applied to image data.
With time series data, the input features are arranged sequentially over time rather than like pixels in an image.
The \textit{tf-keras-vis} implementations can also be used for time series data, assuming the input data has the appropriate format. The gradient-based approaches also work
with $1D$ convolutional layers along the temporal dimension of the input. Thus, saliency maps can be generated for a time series classification model analogous to the standard
case with image data.
Finally, each of the methods receives the normalized time series values $V' \in \mathbb{R}^n$, the trained model $M$, and an optional prediction $y$ (default is the
``best guess'', \ie $y = \operatorname*{argmax}_i P(i \thinspace | \thinspace V') \thinspace \forall \thinspace i \in \{0, 1\}$) as input, and outputs a heatmap
$H \in [0, 1]^n$ highlighting the parts that are most relevant for the classification.
We are interested in values $h \in H$ close to $1$, these are the most important parts of the signal $V'$.
Each value $h_i \in H$ rates the importance of a corresponding input value $v_i \in V', \thinspace \forall \thinspace i \in \{1, \dots n\}$.

\begin{figure*}[t]
    \centering
    \includegraphics[width=\textwidth]{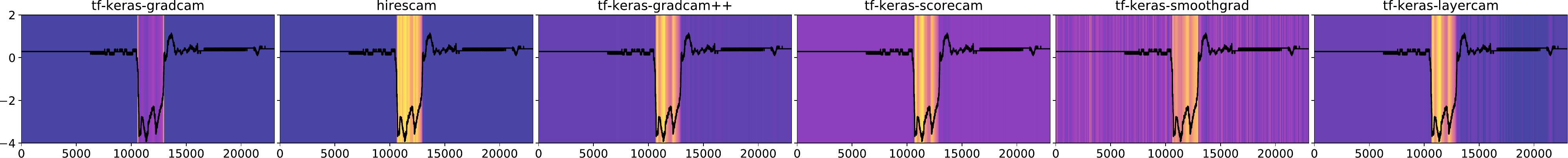}
\caption{Visualization~\cite{bohne:2023}: Heatmap generation methods, \cf~\cite{bohne:2023} for a more detailed discussion; the x-axis shows the respective sampling points, the y-axis the normalized voltage, see Figure~\ref{fig:example:signal:architecture}.}
\label{fig:cam_side_by_side}
\end{figure*}

Since the respective methods have different advantages and weaknesses, the best-suited one depends on the considered task.
Probably most important is accuracy, \ie how well the heatmap reflects the significant regions. Interpretability is also a relevant criterion, as is computational efficiency
and flexibility. The latter refers to the applicability to different types of model architectures.
Figure~\ref{fig:cam_side_by_side} shows a side-by-side plot of all generated heatmaps.
In this example, the different methods agree very well on the relevant regions for the prediction.
In the end, it enables domain experts to assess whether these areas are plausible bases for decision making and allows for knowledge discovery through the resulting KG entries.
Since plausibility may also depend on the context of the classification, the interpretable symbolic state transitions etc. are crucial for the assessment.

If the classification of a whole range of signals reveals that, for instance, certain highlighted segments correlate with certain errors, this is again very valuable knowledge that is
unavailable a priori. For experts, it is often particularly difficult to precisely specify the properties of an ROI. It is often a very intuitive, experience-based and hard
to grasp process, so it could be very valuable to simply learn the ROIs in this way and then compare them to the intuitive notions of human experts. There may also be patterns that
are very subtle and difficult for humans to recognize. Thus, the classification, \ie the neural part of the neuro-symbolic system, benefits from the KG, which essentially narrows
down the search space, and the KG in turn benefits from the neural part, namely from the results and explanation of the ANN-based classification. In the end, if a threshold is
exceeded for an error, \ie a certain number of roughly agreeing heatmaps has been gathered for it, one could crop this sub-ROI and train a classification model for it.
Once this error occurs, the sub-ROI is cropped and the more specialized model is applied, resulting in a sub-ROI patch classification. In conclusion, the system theoretically gets
better at diagnosing errors that it has seen frequently in the past. A further option is to cluster all recorded heatmaps, irrespective of the particular fault, in order to find
patterns. This is only one illustration of the opportunities for knowledge discovery.

\subsection{Neuro-Symbolic Fault Diagnosis}
\label{sec:nesy_fault_diag}

The neuro-symbolic fault diagnosis system proposed in \cite{bohne:2023} makes use of both KGs and their reasoning capabilities as well as explainable learning through CNNs. It thus combines the
advantages of KGs, which incorporate human expertise into the diagnostic process, with the advantages of deep neural networks, which excel at recognizing patterns in complex signals.

The fault diagnosis system is applicable in a variety of domains in which the following conditions are met:
The system to be diagnosed consists of multiple components with causal relationships between one another, \ie an anomaly in one component can transfer to another component, and these causal relationships can be formally
described in a KG. Moreover, each component can be inspected individually, and checking all components would not be an efficient solution, \ie it is desired to minimize the number of inspections. Lastly, it is expected
that (sensor) data like time series or images can be recorded at the components and classified by a neural network.\footnote{Recording tabular data would also be an option, but the chosen neural network architecture and
XAI techniques would have to be adapted accordingly.}

According to the taxonomy in \cite{kautz2022third} that categorizes neuro-symbolic systems based on the integration of the neural and symbolic components, this approach belongs to the \emph{Symbolic[Neuro]} category.
The KG initiates the anomaly classification networks as needed and uses the classification result to guide the subsequent diagnostic step(s).
\begin{figure*}[t]
    \centering
    \includegraphics[width=\textwidth]{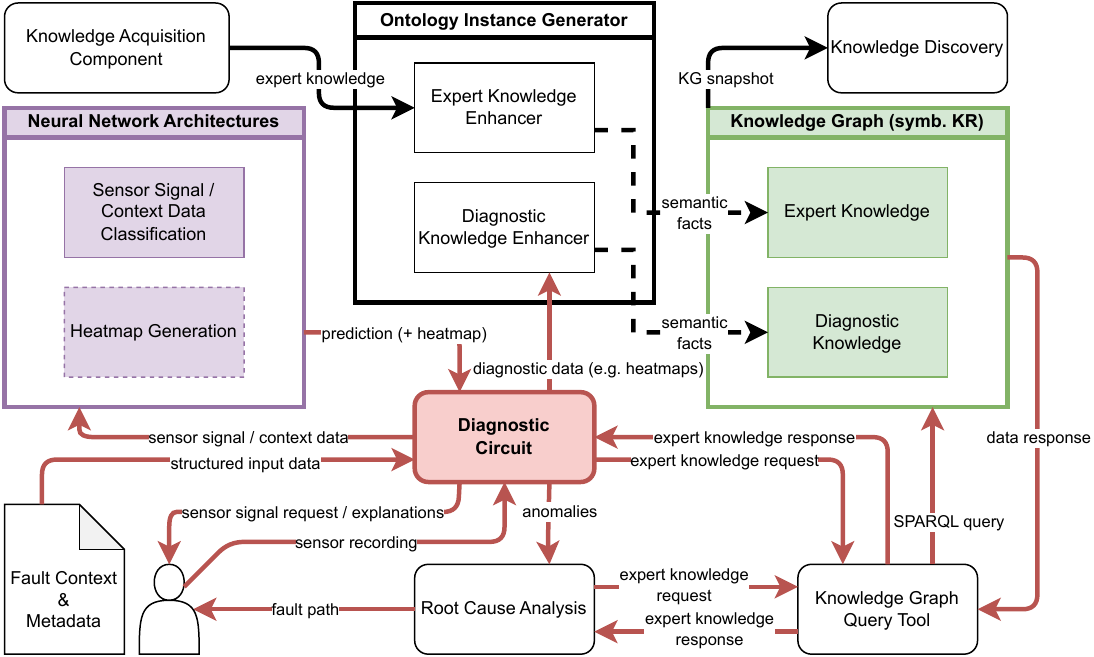}
\caption{Overview (adapted from~\cite{bohne:2023}) of the \textsc{\textcolor{neuro_color}{Neuro}-\textcolor{symbolic_color}{Symbolic}} architecture.}
\label{fig:nesy_architecture}
\end{figure*}

The architecture of the diagnosis system with its various elements and their connections is shown in Figure~\ref{fig:nesy_architecture}.
A central element of the diagnosis process is the diagnostic circuit. It connects the symbolic and neural parts of the system and navigates the diagnostic process. Note that the system is not a continuous monitoring
system, but is only applied on demand if there are indications of an error, \eg through the occurrence of symptoms. Hence, the diagnosis process is started by supplying a fault context and metadata (\eg error codes or
observed symptoms, depending on the domain). Via the \emph{Knowledge Graph Query Tool}, which is a library of predefined SPARQL queries that allow for the retrieval of relevant knowledge from the KG, the diagnostic circuit
accesses expert knowledge that instructs the user to take measurements at a certain component. This instruction is transferred to the user via the diagnostic circuit, who then performs measurements, sensor recordings or
inspections according to the instruction. The measurement is transferred to the neural component of the system, which then classifies it (\eg using a CNN), leading to a binary classification result and optionally to a
heatmap generated through CAM methods, highlighting the most relevant segments of the time series or image for the classification. If the classification result is an anomaly, a \emph{Root Cause Analysis} (RCA) is performed.
An anomaly does not mean that the respective component is defective, as errors can propagate through a chain of components that are causally related. The first component in the chain from which the error originates is
expected to be the root cause. In order to find the root cause, a request is sent via the \emph{Knowledge Graph Query Tool} to the KG, which returns a list of all components that affect the identified anomalous component.
The classification process is repeated recursively until one or more anomalous components are found that are not affected by any other anomalous component, \ie that are root causes. The resulting fault path, which
represents the chain of anomalous components through which the error propagated, is generated and presented to the user, as are the heatmaps at each classification step. This demonstrates the interpretability and
transparency of the diagnosis system which, combined with the explainability of the neural classification methods achieved through CAM approaches, enhances the trustworthiness of the system by allowing the user to
understand and verify the results as well as each step of the process.
At the end of the process, all steps and artifacts (measurements, classification predictions, heatmaps, fault paths, etc.) of the diagnosis process are automatically saved in the KG via the \emph{Diagnostic Knowledge
Enhancer}. The diagnostic knowledge that accumulates over time enables \emph{Knowledge Discovery}, \ie the discovery of previously unknown statistical relationships between certain error patterns and root causes,
thus offering the opportunity of improving the RCA process and making it even more efficient in the future.

The expert knowledge that is a prerequisite for the process needs to be made available by domain experts. This is facilitated by a \emph{Knowledge Acquisition Component}, \eg in the form of
a web interface, in order to ensure that the knowledge can be entered in a structured way before it is added to the KG via the \emph{Expert Knowledge Enhancer}. Crucial knowledge for the functioning of the system are the
causal connections between components and the connections between fault contexts and the suspected components.
The ontology can be adapted to contain many additional domain-specific concepts and relations that are helpful for diagnosis within the respective domain. The following section illustrates an adaptation to the automotive domain.

\subsection{Use Case: Neuro-Symbolic Fault Diagnosis in the Automotive Domain}
\label{sec:instantiation}

Neuro-symbolic fault diagnosis for automobiles, instantiating the approach described in Section~\ref{sec:nesy_fault_diag}, has the potential to relieve mechatronics engineers of many time-consuming
and error-prone tasks. The practical motivation of \cite{bohne:2023} is to tackle the increasing complexity and diversity of modern vehicles, which pose a major challenge
for their diagnosis, \ie to extract meaning from multimodal data that is unmanageable for humans, and to recognize complex patterns.

\paragraph{Symbolic Knowledge Representation}

\begin{figure*}[htb]
    \centering
    \includegraphics[width=\textwidth]{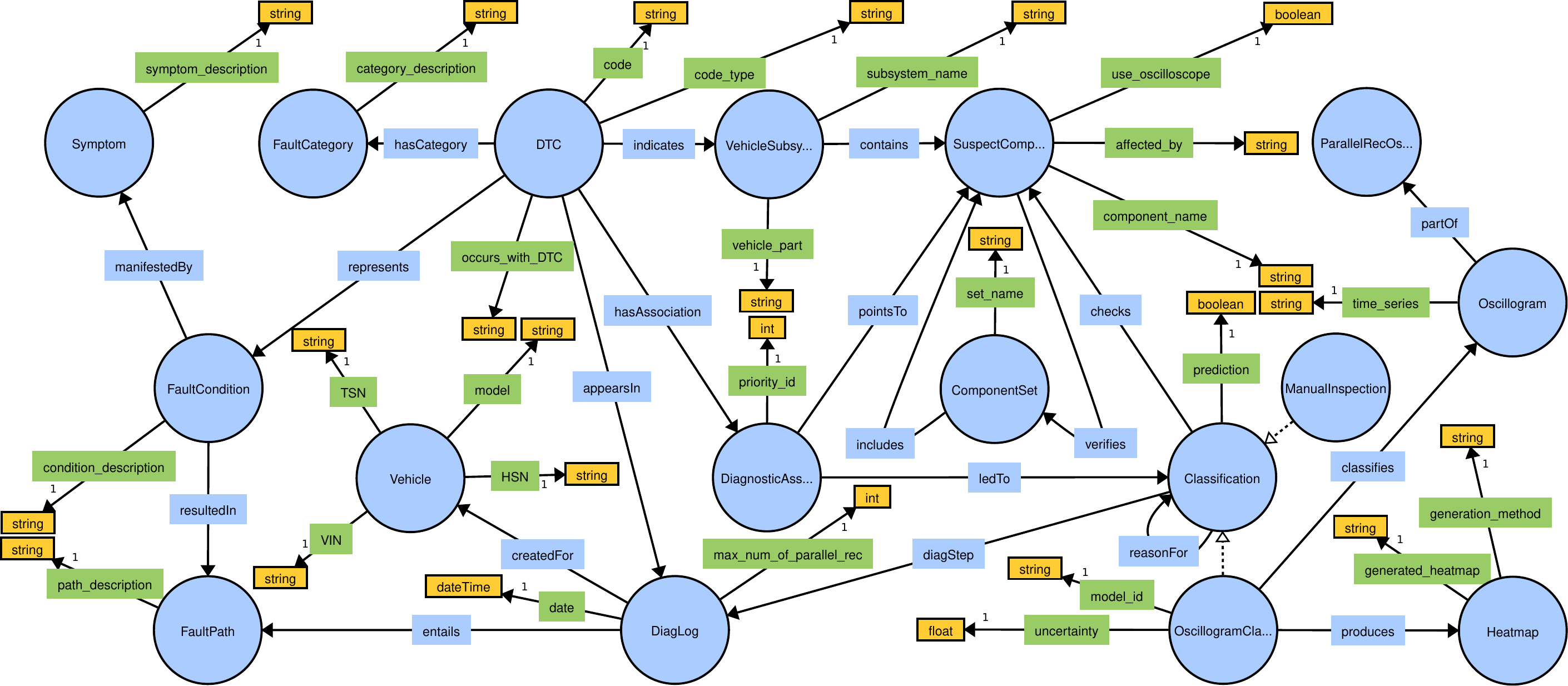}
    \caption{Ontology~\cite{bohne:2023} for capturing knowledge about On-Board Diagnostics (OBD).}
    \label{fig:obd_ontology}
\end{figure*}

To capture and structure diagnosis-relevant knowledge, an ontology\footnote{https://github.com/tbohne/obd\_ontology/releases/tag/v0.1.1} was defined (cf. Figure~\ref{fig:obd_ontology}),
which leads to a KG by populating it with large amounts of instance data, \eg through expert interviews, supplemented with industry partner data. Essentially, there are
three levels of abstraction: The raw definition of the ontology, vehicle-agnostic \emph{expert knowledge} regarding on-board diagnostics (OBD,
ISO 15031-6\footnote{https://www.iso.org/standard/66369.html}), and vehicle-specific \emph{diagnostic knowledge} automatically generated based
on OBD logs read in workshops and acquired as part of the diagnostic process (recorded sensor data, interpretations, etc.).
However, as illustrated in Figure~\ref{fig:nesy_architecture}, the two types of knowledge are not isolated from each other, but connected by meaningful links (\eg connecting
classification instances to the diagnostic associations that led to them) to learn from previous diagnostic runs. All three levels combined constitute the KG. The acquisition of
expert knowledge is accomplished via a web interface (collaborative \emph{knowledge acquisition component}) through which the knowledge is entered, stored in the \textit{Resource
Description Framework} (RDF) format, and hosted on an \textit{Apache Jena Fuseki}\footnote{https://jena.apache.org/documentation/fuseki2/} server. In addition, this knowledge is retrievable in the diagnostic process via predefined SPARQL queries (\emph{KG query tool}), as well as making the KG
expandable and editable in general (\emph{ontology instance generator}).

\paragraph{Expert Knowledge Modeled in the Ontology}

\begin{figure}[htb]
    \centering
    \includegraphics[width=0.66\textwidth]{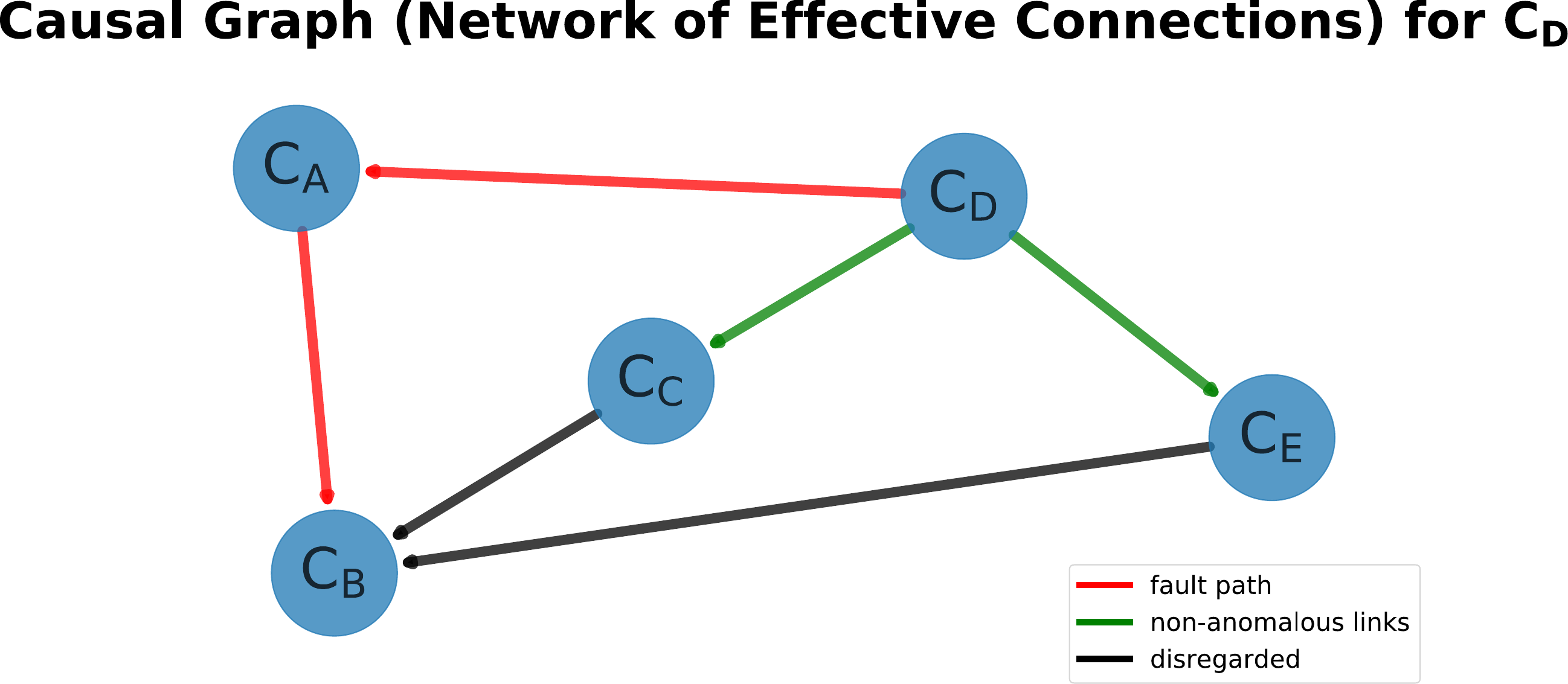}
    \caption{Example~\cite{bohne:2023}: Fault Isolation Result using the Causal Graph.}
    \label{fig:fault_isolation}
\end{figure}

At the core of the knowledge captured in the ontology are the standardized \textit{Diagnostic Trouble Codes} (DTCs), which are a perfect example of what is meant by
\emph{fault context} in the neuro-symbolic architecture visualization in Figure~\ref{fig:nesy_architecture}. In order to efficiently handle the occurrence of DTCs in a vehicle's
electronic control unit (ECU), a DTC parser\footnote{https://github.com/tbohne/dtc\_parser/releases/tag/v0.1} has been developed that is capable of resolving all digits of the
read code and outputting the resulting error information.
A \textcolor{darkgray}{\texttt{\textbf{DTC}}} can have \textcolor{darkgray}{\texttt{\textbf{DiagnosticAssociation}}}s with physical components in a car
(\textcolor{darkgray}{\texttt{\textbf{SuspectComponent}}}). A crucial aspect of such an association is the \textcolor{darkgray}{\texttt{\textbf{priority\_id}}},
based on which components are suggested to be examined in a certain order in the presence of a given DTC. In addition,
\textcolor{darkgray}{\texttt{\textbf{affected\_by}}} represents a list of
other components whose malfunction could affect the correct functionality of the considered component (dependencies can be conceived as a tree, cf. Figure~\ref{fig:fault_isolation}).
Experts can define a \textcolor{darkgray}{\texttt{\textbf{ComponentSet}}} to reduce the number of redundant
diagnostic steps in case there is a specific component that can be leveraged to verify the correct functioning of a whole set of components. Furthermore, each component is contained
in a \textcolor{darkgray}{\texttt{\textbf{VehicleSubsystem}}}, which is associated with a specific part of the vehicle.
Moreover, each DTC represents a \textcolor{darkgray}{\texttt{\textbf{FaultCondition}}} manifested by one or more \textcolor{darkgray}{\texttt{\textbf{Symptom}}}s.

\paragraph{Diagnosis Knowledge Modeled in the Ontology}

There is another theme to the ontology, which is the acquisition and reasonable arrangement of diagnostic data. For each \textcolor{darkgray}{\texttt{\textbf{Vehicle}}}
instance that is entered into the KG, \ie for each vehicle that is diagnosed with the system, a \textcolor{darkgray}{\texttt{\textbf{DiagLog}}} is created that provides the KG with
a kind of explanatory summary of the entire diagnostic process. However, this is not a mere summary, but each entry is automatically sorted into the existing web of expert knowledge
and past diagnostic data by instantiating the concepts of the ontology. Initially, any recorded DTC appears in this log, as this is always the starting point for a diagnosis.
Perhaps most significant are the diagnostic steps, which are also part of the log in the form of \textcolor{darkgray}{\texttt{\textbf{Classification}}} instances that store
their reason, either another classification that detected an anomaly (\textcolor{darkgray}{\texttt{\textbf{reasonFor}}}) or a diagnostic association with a DTC recorded in the
vehicle (\textcolor{darkgray}{\texttt{\textbf{ledTo}}}). Each classification has a binary result (\textcolor{darkgray}{\texttt{\textbf{prediction}}}) that indicates whether the
checked component has an anomaly or not. The concept \textcolor{darkgray}{\texttt{\textbf{Classification}}} has two sub-concepts:
\textcolor{darkgray}{\texttt{\textbf{ManualInspection}}} is a classification performed manually by a mechanic. This is necessary
in cases where oscilloscope-based analysis is not reasonable for a component, \ie \textcolor{darkgray}{\texttt{\textbf{use\_oscilloscope}}} $\coloneqq$
\textcolor{darkgray}{\texttt{\textbf{false}}}, or when we simply cannot provide a classification model
for the specific component yet. The other sub-concept is \textcolor{darkgray}{\texttt{\textbf{OscillogramClassification}}}, which classifies an oscilloscope signal using a
classification model. In this case, we specify an uncertainty value and an ID of the model that produced the classification.
The subject of an oscillogram classification are the \textcolor{darkgray}{\texttt{\textbf{Oscillogram}}}s (time series), which are also stored in the KG and possibly grouped as
an instance of a \textcolor{darkgray}{\texttt{\textbf{ParallelRecOscillogramSet}}}. Finally, we also provide heatmaps for the classification of the oscillograms, which
allow an interpretation of the predictions (cf. Sec. \ref{sec:cam}). \textcolor{darkgray}{\texttt{\textbf{Heatmap}}}s are stored in the KG along with their generation method. The
final diagnosis takes the form of a series of \textcolor{darkgray}{\texttt{\textbf{FaultPath}}}s that start with one component (root cause) and subsequently cascade to others. These fault
paths are not only stored in the diagnostic log, but also associated with the fault conditions, which offers interesting potential for future analysis.

\paragraph{Enhancement of Expert Knowledge}

The \emph{expert knowledge enhancer} can be used to augment the KG hosted by the \textit{Fuseki} server with vehicle-agnostic OBD knowledge. In particular, it generates semantic
facts based on the information entered through a web interface and connects these facts in a meaningful way to what is already available in the KG, \ie it serves as a backend
for the \emph{knowledge acquisition component}. There are three general types of instance data that can be entered along with their specific associated information:
\emph{DTCs}, \emph{vehicle components}, and semantically meaningful \emph{sets of components} grouped together. Alternatively, it is possible to select an existing instance,
view the currently available data, and refine it. Entering a new instance leads to a series of operations in the backend. Quite a number of semantic facts have to be generated
when an expert enters few information. Thus, it is always expert knowledge input via the web interface and corresponding generation of semantic facts in the backend.
All of this is accompanied by a series of input validation mechanisms. This way, a simple KG extension for the expert goes hand in hand with an automatic proper ``wiring'' of
semantic facts in the background.

\paragraph{Enhancement of Vehicle-Specific Diagnosis Knowledge}

As a further element, the \emph{diagnostic knowledge enhancer}, on the other hand, enhances the KG with diagnosis-specific instance data, \ie it connects the OBD data recorded in a particular vehicle,
as well as sensor readings, classifications, etc. generated during the diagnostic process, with corresponding background knowledge stored in the KG. The process typically starts
by creating an instance of the vehicle to be diagnosed, \eg through \textcolor{darkgray}{\texttt{\textbf{extend\_kg\_with\_vehicle("DummyVehicle", "ID2342713")}}}.
Likewise, it typically ends with a call to \textcolor{darkgray}{\texttt{\textbf{extend\_kg\_with\_diag\_log}}}, which takes numerous arguments, including the DTC instances
that are part of the diagnosis and the vehicle identification number (VIN), \eg \textcolor{darkgray}{\texttt{\textbf{["P2563"]}}} and
\textcolor{darkgray}{\texttt{\textbf{"ID2342713"}}}.
This leads, for instance, to \textcolor{darkgray}{\texttt{\textbf{FaultCondition}}} ``\textit{Boost Control Position Sensor Circuit: Implausible Signal}'', represented by
\textcolor{darkgray}{\texttt{\textbf{"P2563"}}}, \textcolor{darkgray}{\texttt{\textbf{resultedIn}}} a \textcolor{darkgray}{\texttt{\textbf{FaultPath}}}, entailed by a
\textcolor{darkgray}{\texttt{\textbf{DiagLog}}} instance, \textcolor{darkgray}{\texttt{\textbf{createdFor}}} the \textcolor{darkgray}{\texttt{\textbf{Vehicle}}} instance with
\textcolor{darkgray}{\texttt{\textbf{VIN}}} \textcolor{darkgray}{\texttt{\textbf{"ID2342713"}}}. Moreover, \textcolor{darkgray}{\texttt{\textbf{"P2563"}}}
\textcolor{darkgray}{\texttt{\textbf{appearsIn}}} this particular instance of \textcolor{darkgray}{\texttt{\textbf{DiagLog}}}. These are only a few examples of the information
collected during diagnosis and its interrelationships. The full set of concepts and relations can be seen in the visualization of the ontology (cf. Figure~\ref{fig:obd_ontology}),
for all of which automatic semantic fact generation and thus KG extension is implemented.

\paragraph{Knowledge Graph Query Tool}

There is a library of predefined SPARQL queries and response processing to access information stored in the KG that is used in the diagnostic process, such as \textcolor{darkgray}{\texttt{\textbf{query\_vehicle\_instance\_by\_vin(vin)}}},
\textcolor{darkgray}{\texttt{\textbf{query\_symptoms\_by\_dtc(dtc)}}} and \textcolor{darkgray}{\texttt{\textbf{query\_suspect\_components}}}\newline
\textcolor{darkgray}{\texttt{\textbf{\_by\_dtc(dtc)}}}. The latter, for instance,
automatically sends and processes the query for a given DTC \textcolor{darkgray}, \eg {\texttt{\textbf{"P2563"}}} in the example~\cite{bohne:2023} shown in Figure~\ref{fig:sample_q}:

\begin{figure}[htb]
    \centering
    \includegraphics[width=0.85\textwidth]{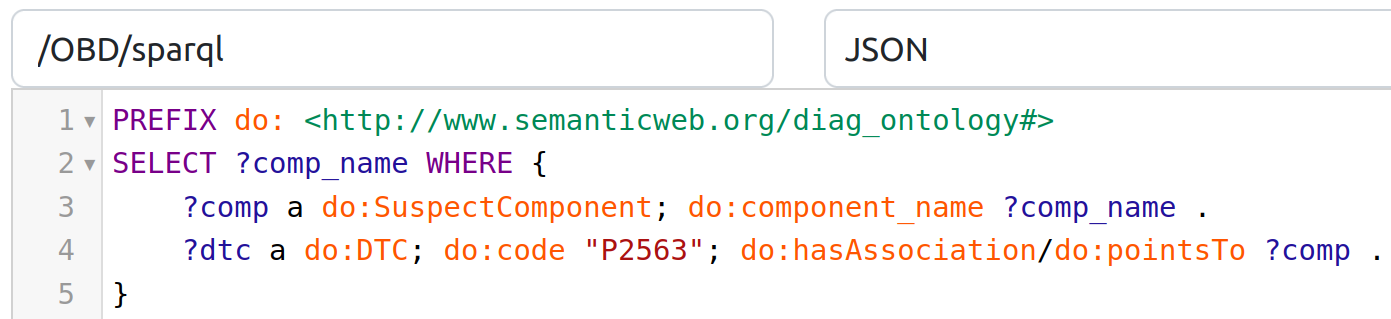}
    \caption[]{Example~\cite{bohne:2023}: SPARQL query to retrieve components for DTC "P2563".}
    \label{fig:sample_q}
\end{figure}

\paragraph{ANN-Based Oscillogram Classification}

\begin{figure}[H]
    \centering
    \begin{subfigure}{0.49\columnwidth}
        \centering
        \includegraphics[width=\linewidth]{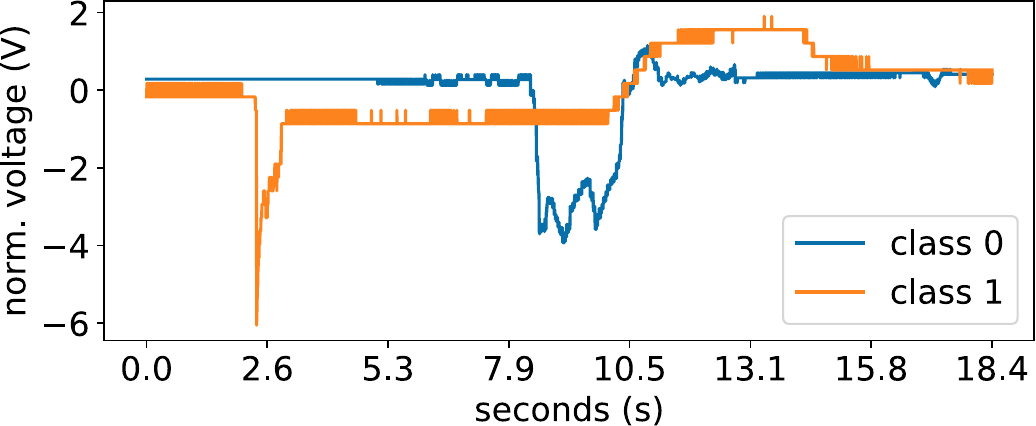}
        \caption{Normalized Battery Voltage~\cite{bohne:2023}.}
        \label{fig:battery_samples}
    \end{subfigure}
    \hfill
    \begin{subfigure}{0.49\columnwidth}
        \centering
        \includegraphics[width=\linewidth]{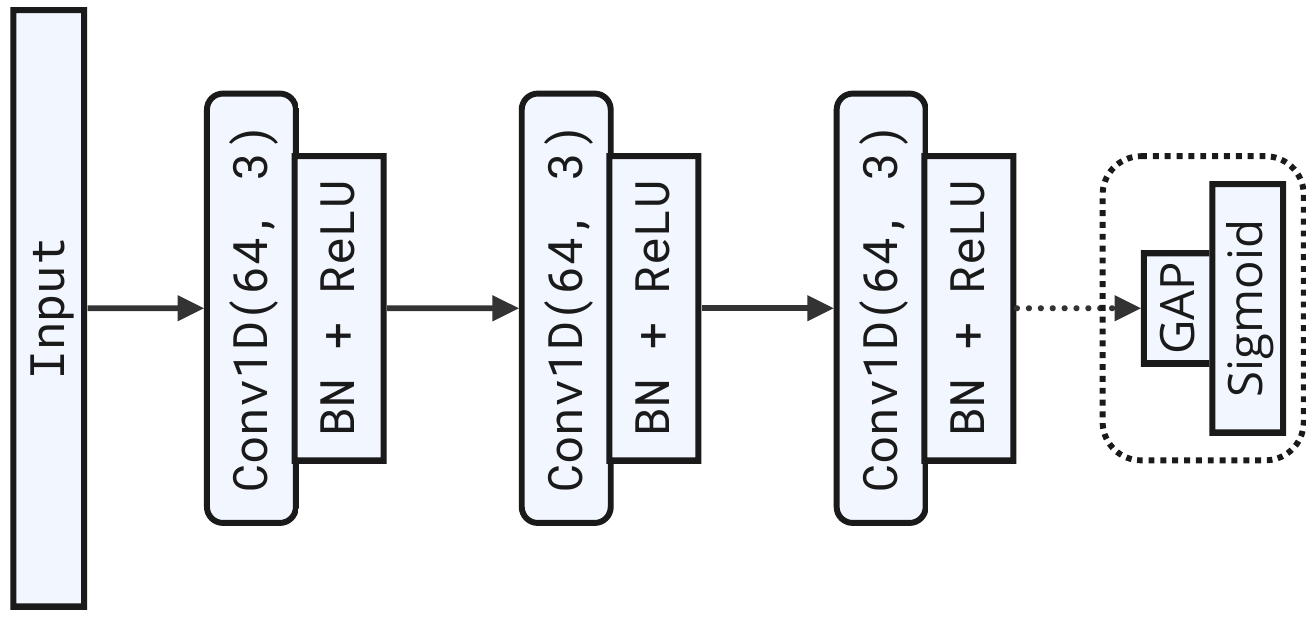}
        \caption{FCN Architecture~\cite{bohne:2023}.}
        \label{fig:fcn}
    \end{subfigure}
\caption{Time series classification and the applied ANN architecture, \cf~\cite{bohne:2023} for more details.}\label{fig:example:signal:architecture}
\end{figure}

A key idea of the developed diagnosis system revolves around sensor information of a certain type. Oscilloscope recordings are performed on specific physical components in the
vehicle to detect indications of problems (anomalies). The recorded oscillograms are fed into a classification model previously trained on a large dataset, which evaluates whether
each recording contains anomalies.\footnote{https://github.com/tbohne/oscillogram\_classification/releases/tag/v0.1.1} The task comes down to binary univariate time series
classification. In the following example, we consider the battery voltage during the engine starting process.
For battery voltage records $V \in \mathbb{R}^n$, performance was best when $z$-normalization was applied to the raw time series data, \ie
$V' := \{\frac{x_i - \mu{V}}{\sigma{V}} \thinspace | \thinspace x_i \in V\}$.
Figure~\ref{fig:battery_samples} shows a regular (1) and an anomalous (0) $z$-normalized voltage sample.
Since the main focus of \cite{bohne:2023} is not to propose a novel ANN architecture for binary classification (anomaly detection) of time series data, we compared several
standard architectures from the literature, made slight adjustments, and selected the best performing one for our purposes, which was a \emph{Fully Convolutional Network} (FCN).
The FCN model (vis. in Figure~\ref{fig:fcn}) is based on \cite{Wang:2017}, in which the authors propose a strong baseline architecture for time series classification.

\paragraph{Neuro-Symbolic Anomaly Detection and Fault Diagnosis}

As the previous sections have shown, knowledge- and machine-learning-based vehicle diagnosis requires the integration of various components.
For the prototypical overall process (\emph{diag. circuit} in Figure~\ref{fig:nesy_architecture}) and to integrate all developed modules, a hierarchical state machine was
defined.\footnote{https://github.com/tbohne/vehicle\_diag\_smach/releases/tag/v0.1.1}, see~\cite{bohne:2023} for a detailed discussion.

Initially, there is meta and context data processing. Based on the read information, the vehicle-specific instance data is entered into the KG. If DTC data is available,
the KG is extended with the processed OBD data, \ie the information that the fault conditions represented by the individual DTCs occurred in the respective vehicle, etc.
If the vehicle instance already exists in the KG, it is extended, otherwise it is newly created. Based on the acquired fault context, the actual diagnostic process is
initiated via the embedded state machine \textcolor{darkgray}{\texttt{\textbf{DIAGNOSIS}}}.

This state machine starts with a state in which a best-suited DTC instance is selected for further processing.
There is a number of possible transitions. If an instance is selected or generated, the process continues with suggesting suspect components. Otherwise, no anomaly was
detected and the indirect conclusion of a potential sensor malfunction is provided. This conclusion should be verified or refuted by the mechanic. Then, either the sensor works, which means that the diagnosis was unsuccessful (only the disproved initial hypothesis and the context are provided due to unmanageable
uncertainty), or there is a diagnosis of a defective sensor. The more interesting case: Certain components in the vehicle are recommended to be investigated in light of the available
information (fault context). Based on the selected DTC instance, the \emph{KG query tool} is used to query the corresponding suspect components, and for each, whether it can be
reasonably diagnosed with an oscilloscope. Afterwards, we first distinguish between the subset of suspect components for which oscilloscope diagnosis is appropriate and those that
must be verified manually. Then, synchronized \emph{sensor recordings} are performed at the proposed components of the respective subset, and the resulting \emph{oscillograms are
classified} using trained ANN models. The prediction can be interpreted by \emph{overlaid heatmaps} (cf. Sec. \ref{sec:cam}). Subsequently, the subset of recordings to be inspected
manually is handed over to the mechanic. In the end, there is a set of anomalous components identified by the trained models and the mechanic. If this set is empty, \ie no anomaly
has been detected, then the next iteration of suggestions follows. However, if no anomaly is detected and there are no remaining components to suggest, then the next DTC instance is selected.
If anomalies are found, though, the \emph{root cause analysis} in Figure~\ref{fig:nesy_architecture} follows. For further details on the diagnostic procedure, we refer to~\cite{bohne:2023}.

\paragraph{Root Cause Analysis to Determine the Source of the Defect}

Once an anomaly is identified in the described manner, the fault is isolated by recursively inspecting the cause-effect relationships in the vehicle, which are part of the KG
(cf. Figure~\ref{fig:obd_ontology}), \ie graph traversal coupled with anomaly detection. This basically creates a causal sub-graph for each anomalous component from which the
root cause can be derived. After all, in technological systems errors are rarely encountered that are entirely independent of other components in the system. Typically, there
are cascading paths where a problem starts at one component and then spreads to others. The actual interest is directed at the root cause instead of mere side conditions.
The termination criterion is that there are no further known components that have not yet been examined and that could directly affect an anomalous component. During the
recursive procedure, the fault path, explicitly considered links, etc. are tracked. The result is visualized dynamically for each diagnosis, \eg Figure~\ref{fig:fault_isolation}.
The recursive and dynamic sub-graph construction is based on KG queries
using the initial anomalous component $C_D$.
After isolating the problem, the diagnosis is entered into the KG, along with a detailed record of all relevant information that led
to it, in order to learn from it and facilitate future diagnoses.

Ultimately, the diagnosis is presented in the form of the inverted fault path in Figure~\ref{fig:fault_isolation} as this was the \textit{affected-by}-direction, not the direction
starting from the probable root cause of the fault. For the example in Figure~\ref{fig:fault_isolation}, this would be $\{C_B \rightarrow C_A \rightarrow C_D\}$. So the problem
probably started at $C_B$, cascaded through $C_A$, and finally to $C_D$. Although it is not guaranteed to be the actual root cause, it should provide an experienced mechanic with
a fairly strong understanding of the problem prevailing in the vehicle in question.

\section{Conclusions}\label{sec:conclusions}

In this chapter, we provided an overview on knowledge-augmented, explainable and interpretable learning in the context of anomaly detection and diagnosis. For this, we have outlined the general context and discussed exemplary methods in this domain, specifically considering the articles~\cite{AS:17,SA:18} for discussing pattern mining in the context of anomaly detection as well as~\cite{ABP:05Score,BAKP:06,ABP:06} on learning scoring systems. Furthermore, we provided a significantly adapted summary of the article~\cite{bohne:2023} -- on a neuro-symbolic approach for anomaly detection and diagnosis. Thus, we covered simple knowledge representations such as diagnostic scores, pattern-based methods in combination with various forms of symbolic knowledge, as well as a hybrid neuro-symbolic method integrating deep learning with symbolic knowledge.

For future work, further extending knowledge-augmented approaches, particularly regarding advanced neuro-symbolic approaches, is a major interesting research direction. Here, directly integrating interpretability and explainability in the methodology is an interesting option to consider. Specifically, this also relates to different types of complex data, \eg graphs, images, etc.
Moreover, we are planning to work on a further generalization of the neuro-symbolic diagnosis framework presented in~\cite{bohne:2023} as well as a systematic evaluation of its architecture using randomized, parameterized and
domain-agnostic synthetic problem instances and corresponding ground truth solutions generated based on the abstract formulation of the general problem of diagnosing systems with interconnected components based on sensor
signal assessment.

\end{document}